_______________________________________________________________________________________

# Optimization of Transfer Learning for Sign Language Recognition Targeting Mobile Platform

Dhruv Rathi

Department of Computer Engineering

Delhi Technological University

New Delhi, India

*Abstract*—The target of this research is to experiment, iterate and recommend a system that is successful in recognition of American Sign Language(ASL). It is a challenging as well as an interesting problem that if solved will bring a leap in social and technological aspects alike. In this paper, we propose a real-time recognizer of ASL based on a mobile platform, so that it will have more accessibility and provides an ease of use. The technique implemented is Transfer Learning of new data of Hand gestures for alphabets in ASL to be modelled on various pre-trained high- end models and optimize the best model to run on a mobile platform considering the various limitations of the same during optimization. The data used consists of 27,455 images of 24 alphabets of ASL. The optimized model when ran over a memory-efficient mobile application, provides an accuracy of 95.03% of accurate recognition with an average recognition time of 2.42 seconds. This method ensures considerable discrimination in accuracy and recognition time than the previous research.

_______________________________________________________*****________________________________________________________

## I. INTRODUCTION

The aimofdeveloping a real-time working system to accurately read the sign language and communicate with the signer in an automated sense, without having an expertise or even an introductory knowledge in that form of communication has been a topic of interest to the researchers of Computer Vision and Natural Language Processing, thus producing some noteworthy works which have had a great impact both in terms of algorithmic advancements and development of society. The problem of Sign Language Interpretation(SLI) is a form of the standard Action Classification Problem when seen from the aspect of Deep Learning. Majority of the previous works that have been able to set a benchmark in the field of SLI are such structured in their implementation that they are computationally expensive enough to be unsuitable for utilizing as a daily life tool limiting themselves through their dependency on hardware or are in limited to statically detecting and analyzing the gestures and sign languages, thus a robust and accurate system is needed.

In this work, the aim is to develop a novel real-time implemented system based on Mobile platforms that recognize and extracts sign actions of the American Sign Language(ASL) from live video frame to frame and construct sentences by monitoring the continuous action movements performed by the signer, i.e., the ASL will be converted to words in real time recording on an Android-based mobile phone. A visual framework for representation is provided that shows the corresponding likelihood of top 3 words and the time taken to for recognition.

The dataset used for the discussed research is Sign Language MNIST - Drop-In Replacement for MNIST for Hand Gesture Recognition Tasks [1]. A detailed discussion of the dataset is done in Section III(A).

To obtain a system that is accurate to the desired level and simultaneously holds the capability to run on a generic hardware deficient Android mobile phone, the system needs to be perfectly optimized to feed and satisfy itself with the low computational prowess. The best suitable approach to proceed with which will meet both the discussed parameters is the application of Transfer Learning. This is due to the possibility of high-optimality of the models and ease of training even with low amount of data as will be discussed in Section III(B).

Our proposed method uses Transfer Learning which makes the process and thus the system more robust with both large or small datasets. Training a powerful Deep Learning model from scratch can take days at a time and requires appreciable computational power and hardware, while Transfer Learning makes those models flexible and they easily adapt to new data. The obtained accuracy of recognition is 95.03%.

The remaining paper is catalogued as follows: Section II. Reviews the previous research done in the field. Section III: A) Delineates the dataset used B) Review and mathematics of Transfer Learning. Section IV: A) Discusses the experiment models and their structures B) Discusses the proposed experimentation processing C) Optimization of retrained model for Mobile Environment. Section V: A) Discusses and evaluates the results obtained from the experimentation B)

_______________________________________________________________________________________

Method implementation and output C) Discusses the future work. Section VI: Mentions the references.

## II. PREVIOUS WORK

There are various approaches that have been proposed by researchers in the form of patents and research paper for the purpose of identification of gestures and signs pertaining to sign language.

G. Anantha Rao et. al. [2] discusses the results of the application of Deep Convolutional Neural Networks for Sign Language Recognition, with an accuracy of 92.88% recognition on a self-constructed dataset using OpenCV and Keras libraries.

Amrutha C U et. al. [3] in the paper "Improving language acquisition in sensory deficit individuals with mobile application" discusses construction of a mobile application that is a system of a speech-based transformer from text to Indian Sign Language (ISL) from a pre-built database of images of signs stored locally on the machine and are queried during runtime.

Sébastien Marcel et. al. [4] in the paper "Hand Posture Recognition in a Body-Face centred space" uses a neural network to interpret hand postures in an image. The dataset was self-constructed containing uniform and complex backgrounds with a recognition accuracy of 93.7% and 84.4% respectively.

## III. DATASET USED AND TRANSFER LEANING

### A. The Dataset Used

The open dataset given at Kaggle called Sign Language MNIST - Drop-In Replacement for MNIST for Hand Gesture Recognition Task [1] which contains set of 28x28 images of all the alphabet, except J and Z, of the standard American Sign Language(ASL). The data contains a total of 27,455 cases. A sampled image set can be observed in Fig. (1)

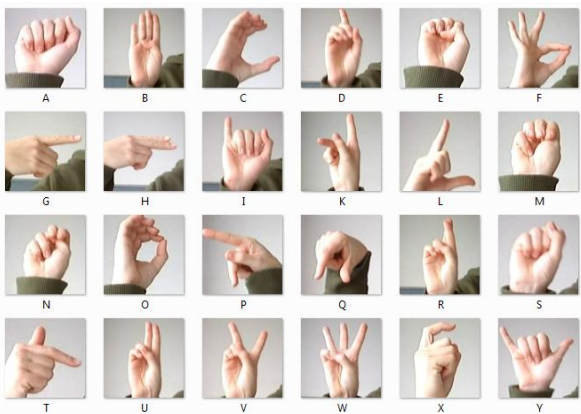

Fig. (1)

The data in its raw form is provided as a pixel to pixel intensity [0-255] class-wise distributed XLS files. The data preprocessing steps included conversion of the mentioned data to image format using Python's open-source libraries Pandas,

Numpy and others to obtain PNG format 28x28 grayscale images.

### B. What is Transfer Learning?

Transfer Learning is an advanced technique of Deep Learning where a model developed for a task is used as a starting point for a model on a second context similar task. Training a Convolutional Neural Network(CNN) on ImageNet [give refer] dataset takes around 2-3 weeks across multiple GPUs. In the current research, two pre-trained models, Inception V3 model [5] and MobileNets [6] open-source models for comparison purposes. Both the mentioned models utilize CNN at their cores as discussed further.

In Transfer Learning technique, the last layers or some of the last layers of the pre-trained models, depending on the data being used are unfrozen and retrained on the new dataset to obtain required results. If only the last layer in the model is unfrozen, the resulting partial model then gives the probabilities of the outputs against each other rather than giving the most probable output, this altering will be much beneficial for sentence construction which will be discussed in the future work. If multiple last layers are unfrozen then a smaller Learning Rate is expected during retraining with the new dataset as this will help preserve the weights of the last few layers.

The MobileNets are small, low-latency, low-power models parametrized to meet the resources constraints of a mobile device. Google Research has 16 pre-trained open-source MobileNet models with varying parameters and accuracy competing for accuracy against time. MobileNets are mobile first Deepwise Separable Convolution Neural Networks, counting depthwise and pointwise convolutions as different layers, MobileNets has 28 layers and is trained on the ImageNet dataset.

The Inception V3 model that is used in the current research is a successor of Inception V1 model which is a variant of GoogleNet[7]with the additional features of Batch Normalization and the introduction of Factorization of larger convolutions into smaller convolutions. The main theory of models with Inception architecture is to find out how an optimal local sparse structure in a convolutional vision network can be approximated and covered by readily available dense components.

## IV. PROPOSED EXPERIMENTATION AND RESULTS

Here we present a method of experimentation that results into a mobile-based engine that successfully recognizes hand gestures and signs pertaining to the alphabets in the American Sign Language (ASL). The research is aimed specifically towards mobile-based system so that the system in real time implementation can be used as an everyday tool and does not suffer from mobility limitations. The first step in the experimentation is the construction of a Deep Learning model that could successfully recognize the alphabet singularly after every requested frame, thus two models, 1) mobilenet_0.50_224 2) Inception V3 are trained using Transfer Learning and further optimized and further re-optimized to obtain a satisfactory model to run on mobile devices. Further, we will see implementation, experiments and results of both the models,

### A. MobileNet_v1_0.50_224 model

The MobileNet_v1_0.50_224 model has 1.24 million parameters. The top layer of mobilenet_v1_0.50_224 takes as input a 1001-dimensional vector for each image, we train a softmax layer on top of this representation. Assume that the softmax layer contains N labels, this corresponds to learning N+1001*N model parameters corresponding to the learned biases and weights. When the process of Transfer Learning is initiated on the mentioned model, an analysis of all images is done and bottlenecks for each are calculated and stored, the bottleneck is referred to as the layer just before the final output layer that is responsible for the classification. This reasoning about why the training of the last layer alone still results in such a high accuracy is that the weights that are stored corresponding to each layer after forward and backward propagations during the long training durations are responsible for the classification of all the 1,000 objects in ImageNet will serve the same purpose with the new classes that we are aiming to classify with our provided data.

Our method as generalized showcases two variants of accuracy, Training accuracy, which is the measure of the percentage of images that were used in the current training batch and were correctly-labelled, another is Validation accuracy, which is the measure of precision on a randomly selected group of images from a varying set. It can be distinctly observed from the graphs in Fig. (4) and Fig. (6) that the validation accuracy fluctuates at irregular intervals, this is due to the reasoning that a random subset of the validation set is chosen for each validation accuracy measurement. Similar to as in Deep Learning training, the hyper-parameters are the ultimate game changers in Transfer Learning, several iterations were performed with varying hyperparameters, which in the present case includes, the Learning Rate (LR), which controls the magnitude of the update to the final layer during training.

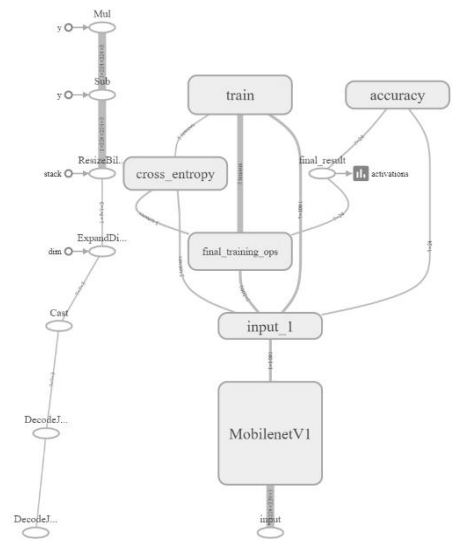

Fig. (2)

The MobileNet_v1_0.5_244 model is retrained on the mentioned dataset with an LR = 0.01 and number of training steps = 5000, with a training batch size of 100 images, from the dataset, the division of data is as follows, 10% data is kept for testing purposes, 10% data is kept for validation purposes and 80% data is used for training. On training the model on the later mentioned hardware, the training accuracy is 97.24% and the validation accuracy is 95.06%. Fig. (2) represents the graph obtained from Tensorboard, a visualization tool provided by Tensorflow [8], that provides an insight into the retrained MobileNet model and its architecture.

### B. Inception V3 Model

The Inception V1 was developed as a variant of GoogleNet containing about 24 million parameters, with further versions V2 and V3, introducing Factorization of convolutions into smaller convolutions and addition of Batch Normalization to the fully connected layer respectively. The architecture of Inception model is a 299x299x3 input representing a field of 299 pixels and 3 channels in the image, combined with five convolutional layers, in addition to a handful of max-pooling operations, and successive stacks of Inception Modules, which are naively a set of different convolution filters and a max-pool filter combined undergoing filter concatenation. The output is a softmax layer at the end. The top layer of Inception V3 model takes as input a 2048-dimensional vector, we train a softmax layer on top of it. Assuming the softmax layer contains N labels, this corresponds to learning N + 2048*N parameters corresponding to learned weights and biases.

The hyper-parameters and accuracy metrics are similar as that of in the MobileNet case.

The Inception model is retrained on the mentioned dataset with an LR = 0.01 and number of training steps = 5000, with a training batch size of 100 images, from the dataset, the division of data is as follows, 10% data is kept for testing purposes, 10% data is kept for validation purposes and 80% data is used for training. On training the model on the later mentioned hardware, the training accuracy is 98.01% and the validation accuracy is 93.36%. Fig. (3) represents the graph obtained from Tensorboard, that provides an insight into the retrained Inception V3 model and its architecture.

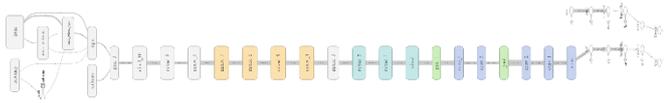

Fig. (3)

### C. Optimization of a retrained model for Mobile Environment

A modern smartphone can run about 10 Giga Floating Point Operations (GFLOPs) per second, thus the best expectant result from a 5 GFLOP model is two frames per second, this model and hardware dependency enables us to run the model even on limited RAM and high latency. Till now we have successfully retrained models that give us satisfactory accuracies for the purpose of Sign Language Recognition of alphabet in ASL, but when running any model on a mobile device, we might face three kinds of issues, 1) The Model and Binary Size; 2) App Speed and model loading speed; 3) Performance and Threading. Tensorflow [8] provides several benchmark and optimization tools that counter the issues above.

The benchmark_model tool gives an estimate of the total number of floating point operations that are required to execute the graph, extrapolating this will lead us to get an idea of the minimum memory requirement for running the model considering in mind a satisfactory User Experience.

Model size is another potential part which can undergo high optimization. Transfer Learning models can be quite large in size given the huge training times and data involved, while a mobile phone has a limited memory that it has and that the operating system can allocate. Thus, the aim is to remove unusable and incompatible parts from the model, the important part of this is the number of constant parameters, let's assume M, stored as 32-bit float. A rough estimate of the size of the model is 4*N. Experimentation shows that only a small cutback in accuracy is found by replacing 32-bit by 8-bit per parameters, but the model size reduces by 75%.

Memory footprint is the most commendable technique to speed up the loading of models in the application. The idea is to load a file using memory mapping, rather than using the usual method of I/O APIs calling, in which an area of memory is allocated on the heap and then copying bytes from disk into it, the model is loaded directly into the memory, once the Operating System is notified through a required call, thus giving the advantage of 1) Increased speed in loading, 2) Reduces Paging, 3) The RAM budget dependency is thus removed from the whole system. Tensorflow handles this functionality with a function LoadMemoryMappedModel().

Thus we obtain an optimized model that is ready to be run on a mobile device, next step is the development of a JAVA based application for Android devices that incorporates at the least the following functionalities, 1) A camera interface that is used to capture the video frames during the act of performing the sign and gestures 2) An environment that is capable to process the developed model successfully, capture and record the results 3) An interface to indicate the results of the processing.

## V. DISCUSSION OF RESULTS AND ITS SIGNIFICANCE

### A. Results and Comparison

Upto this point the whole discussion revolves around experimentation with Inception V3 and MobileNet models, this is done with the aim to find the best model that will suit our needs, for the clarification of the same, we need to test the resulting models in real time environment and need a detailed visualization of the accuracy results of both,

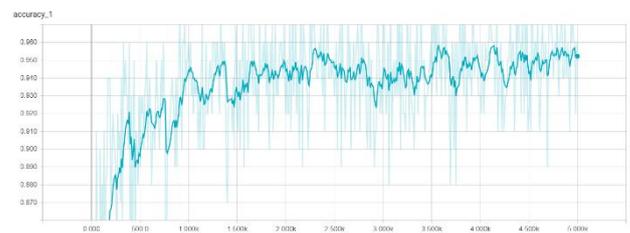

Fig. (4)

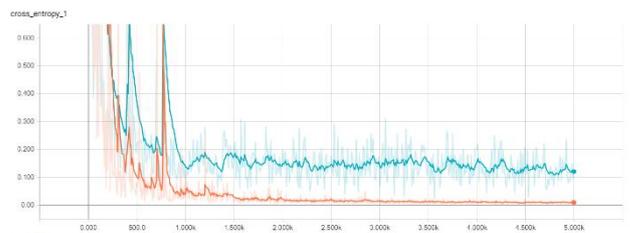

Fig. (5)

Fig. (4) is the Accuracy graph of Validation set of the retrained MobileNet model against training steps (5,000) and Fig. (5) represents the cross-entropy value of the retrained MobileNet model during training of the model, against training steps, colour coding of orange is indictive of Training data and Sky-blue that of Validation data, the final validation accuracy is 95.06%.

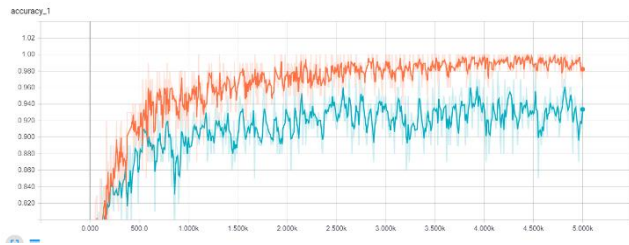

Fig. (6)

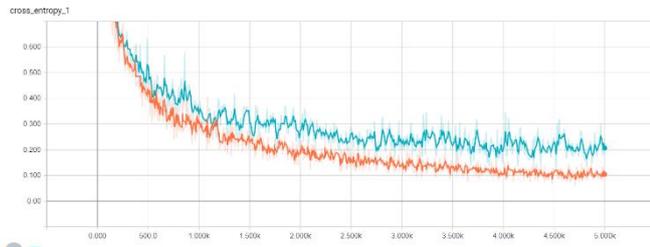

Fig. (7)

Fig. (6) is the Accuracy graph of the Training set and the Validation set of the retrained Inception V3 model against training steps (5,000) and Fig. (7) represents the cross-entropy value of the retrained Inception V3 model during training of the model, against training steps, colour coding of orange is indictive of Training data and Sky-blue that of Validation data, the final validation accuracy is 93.36%. The training accuracy, in this case, is 98.23% which indicates that the results are authentic and there is no instance of over-training.

| S. No. | Model Name | Parameters (in million) | Time to retrain | Accuracy |
|--------|------------|-------------------------|-----------------|----------|
| 1. | MobileNet | 1.24 | 8.51 min. | 95.06 % |
| 2. | Inception_V3 | 24 | 12.31 min. | 93.36% |

Table (1)

From the output graphs, it is evident that MobileNet model outperforms Inception V3, although not by a very wide gap. But, upon implementation and testing the optimized models in a real time environment in a physical mobile device, the observation is that MobileNet outperforms Inception V3 by a very high measure in its accuracy of recognition of sign language alphabet, this statement can be attributed to different model architectures of the two models that will lead to learning of different parameters against the dataset provided and thus giving varied accuracies in recognition during the real time testing, as the images being processed contains numerous factors such as lighting conditions, varied backgrounds and others, which different models adapt to in their unique way.

The training of the models was done on the following configuration, NVIDIA GPU GeForce 940MX, 8GB RAM on Laptop hardware.

*B. Real Time implementation and Output*

An output of the real-time testing of the MobileNet modelled version of the application gives following outputs in recognition of the alphabet A, S and L. The boxes on the top of the screen in the results are indictive of the predicted probability of top three choices by the model.

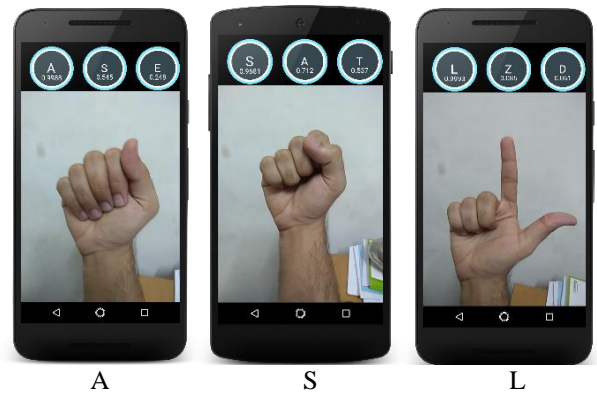

A                    S                    L

Fig. (8)

Fig. (8) shows the ASL alphabet A, S and L with some noise in the background and achieve correct probabilities of 0.9988, 0.9681 and 0.9993 respectively. The other two probabilities for each of the three alphabets are separated by a considerably large difference so as to distinguishably declare the winner output.

*C. Future Work*

The discussed research has been successful in developing a system that can interpret American Sign Language alphabets in real time environment and thus can potentially act as a communication device between the signer and a non-signer and thus will help both technically and socially.

The researcher aims towards continuing the current ideation and system to include Natural Language Processing (NLP) and other techniques for sentence construction.

Also aimed at arethe construction ofproper datasets for other type of Sign Languages and inclusion of gestures and expanding the research by combining the previous point to develop a context-specific system for sentence construction that can ultimately act as a defined language interpreter.

## VI. REFERENCES

[1] https://www.kaggle.com/datamunge/sign-language-mnist/, Sign Language MNIST, Kaggle, 2017.

[2] G.Anantha Rao, K.Syamala , P.V.V.Kishore, A.S.C.S.Sastry, Deep Convolutional Neural Networks for Sign Language Recognition, SPACES-2018, Dept. of ECE, K L Deemed to be UNIVERSITY, 2018.

[3] C. Amrutha C U,Nithya Davis, Samrutha K S, Shilpa N S, Job Chunkath, Improving language acquisition in sensory deficit individuals with, International Conference on Emerging Trends in Engineering, Science and Technology (ICETEST, 2015.

[4] S. Marcel, Hand Posture Recognition in a Body-Face centred space, France Telecom CNET.

[5] Christian Szegedy, Vincent Vanhoucke, Sergey Ioffe, Jonathon Shlens, Rethinking the Inception Architecture for Computer Vision, 2016 IEEE Conference on Computer Vision and Pattern Recognition (CVPR), Las Vegas, NV, 2016, pp. 2818-2826., 2015.

[6] Andrew G. Howard, Menglong Zhu, Bo Chen, Dmitry Kalenichenko, Weijun Wang, Tobias Weyand, Marco Andreetto, Hartwig Adam, MobileNets: Efficient Convolutional Neural Networks for Mobile Vision Applications, https://arxiv.org/abs/1704.04861, 2017.

[7] Christian Szegedy,Wei Liu, Yangqing Jia, Pierre Sermanet, Scott Reed, Dragomir Anguelov, Dumitru Erhan, Vincent Vanhoucke, Andrew Rabinovich, Going Deeper with Convolutions, https://arxiv.org/abs/1409.4842, 2014.

[8] Martín Abadi, Ashish Agarwal, Paul Barham, Eugene Brevdo, Zhifeng Chen, Craig Citro,TensorFlow: Large-Scale Machine Learning on Heterogeneous Distributed Systems, https://arxiv.org/abs/1603.04467, 2016.